# Multiobjective Optimization Analysis for Finding Infrastructure-as-Code Deployment Configurations

ENEKO OSABA*

TECNALIA, Basque Research and Technology Alliance (BRTA), 48160 Derio, Bizkaia, Spain

JOSU DIAZ-DE-ARCAYA

TECNALIA, Basque Research and Technology Alliance (BRTA), 48160 Derio, Bizkaia, Spain

JUNCAL ALONSO

TECNALIA, Basque Research and Technology Alliance (BRTA), 48160 Derio, Bizkaia, Spain

JESUS L. LOBO

TECNALIA, Basque Research and Technology Alliance (BRTA), 48160 Derio, Bizkaia, Spain

GORKA BENGURIA

TECNALIA, Basque Research and Technology Alliance (BRTA), 48160 Derio, Bizkaia, Spain

IÑAKI ETXANIZ

TECNALIA, Basque Research and Technology Alliance (BRTA), 48160 Derio, Bizkaia, Spain

Multiobjective optimization is a hot topic in the artificial intelligence and operations research communities. The design and development of multiobjective methods is a frequent task for researchers and practitioners. As a result of this vibrant activity, a myriad of techniques have been proposed in the literature to date, demonstrating a significant effectiveness for dealing with situations coming from a wide range of real-world areas. This paper is focused on a multiobjective problem related to optimizing Infrastructure-as-Code deployment configurations. The system implemented for solving this problem has been coined as *IaC Optimizer Platform* (IOP). Despite the fact that a prototypical version of the IOP has been introduced in the literature before, a deeper analysis focused on the resolution of the problem is needed, in order to determine which is the most appropriate multiobjective method for embedding in the IOP. The main motivation behind the analysis conducted in this work is to enhance the IOP performance as much as possible. This is a crucial aspect of this system, deeming that it will be deployed in a real environment, as it is being developed as part of a H2020 European project. Going deeper, we resort in this paper to nine different evolutionary computation-based multiobjective algorithms. For assessing the quality of the considered solvers, 12 different problem instances have been generated based on real-world settings. Results obtained by each method after 10 independent runs have been compared using Friedman's non-parametric tests. Findings reached from the tests carried out lad to the creation of a multi-algorithm system, capable of applying different techniques according to the user's needs.

* Corresponding author: eneko.osaba@tecnalia.com.

**ACM Reference Format:**
Eneko Osaba, Josu Diaz-de-Arcaya, Juncal Alonso, Jesus L. Lobo, Gorka Benguria and Iñaki Etxaniz. 2023. Multiobjective Optimization Analysis for Finding Infrastructure-as-Code Deployment Configurations. In ICCCM '23: The 11th International Conference on Computer and Communications Management, August 04-06, 2023, Nagoya, Japan. ACM, New York, NY, USA, 9 pages.

## 1 INTRODUCTION

The correct solving of multiobjective problems is a recurrent task in artificial intelligence and operations research fields. The remarkable popularity of this area is clearly justified because Multi-Objective Problems (MOP) are particularly frequent in real-world environments. In a nutshell, the addressing of this kind of problems involves the searching of a set of solutions that provide the optimal equilibrium among all the objectives at hand [1]. In MOP optimization, a solution can be considered as Pareto optimal if there is no other outcome better than it in all the deemed objectives. For this reason, the objectives that compose a MOP must be competitive among them.

A plethora of solvers has been developed in the last decades for facing MOPs. Methods coming from fields such as Evolutionary Computation (EC, [2, 3]) and Swarm Intelligence (SI, [4, 5]) have been especially resorted in the last years. Some of the most well-known methods are the Nondominated Sorting Genetic Algorithm II (NSGA-II, [6]), or the multiobjective evolutionary algorithm based on decomposition (MOEA/D, [7]). Regarding application fields, MOPs have demonstrated to be effective for addressing different situations in areas such as logistics [8], energy [9], industry [10] or engineering [11].

In this paper, we embrace MOP optimization for dealing with a problem related to optimizing Infrastructure-as-Code (IaC, [12]) deployment configurations. More specifically, this paper gravitates around a system coined as *IaC Optimizer Platform* (IOP), which objective is to find optimized deployment compositions of the IaC on the most appropriate infrastructural elements that fit some preset requirements and objectives. Among the objectives to optimize, some mutually opposed goals can be found, such as the cost of the whole deployment, and its overall performance.

The IOP has been previously introduced in the literature, in the research conducted in [13]. In that article, the system is preliminary introduced, and its application is superficially shown. In that prototypical version, the IOP resorts to NSGA-II for carrying out the optimization. In the present paper, we took a step forward in the development of the IOP, undertaking a detailed experimentation to determine which is the most suitable multiobjective method to deal with the problem of finding optimized IaC deployment configurations.

For performing this experimentation, we test the behavior of nine different multiobjective EC solvers over 12 instances of the problem. These use cases are heterogeneous enough to assure that the conclusions drawn are significant. Details on the techniques used and instances generated are provided in following sections.

The interest on conducting this study is remarkably high since the IOP is expected to work on a real environment. For this reason, its performance should be as good as possible, and the selection of the optimization algorithms employed should be made after a thorough and rigorous study. Finally, it is noteworthy that the IOP has been implemented as part of a Horizon 2020 EU research project. The main goal of this project, named PIACERE (https://www.piacere-project.eu/) is to develop a complete framework for the development, deployment, and operation of IaC running on cloud continuum.

This work is structured as follows. In the following Section 2 we introduce the optimization problem to solve. In Section 3 we focus on describing the IOP and all the algorithms that have been considered for the tests carried out on this work. Section 4 is devoted to the experimentation conducted, while Section 5 finishes this paper with conclusions and further work.

## 2 PROBLEM STATEMENT AND FUNDAMENTALS

To adequately comprehend the optimization problem at hand, let us introduce an example in this section. First, it should be mentioned that, for properly conducting the optimization process, the IOP counts with an Infrastructural Elements Catalog, or IEC, which is full of elements that the solver can consider in order to select the most optimal deployment configuration. Thus, we introduce here a simplified IEC composed of three kinds of elements, [`Storage (ST), Database (DB), Virtual Machine (VM)`], having five options for each of these categories:

- `ST`: [St EU, St Spain, AZ 2, G 2, St Spain, St USA]
- `DB`: [mysq.GB, db.dyn, postgreSQL.GR, r4.large, m3.med]
- `VM`: [C1France, C2Europe, m5.large, t2.nano, DS13v2]

Additionally, all elements have some associated attributes, such as the expected performances, the overall availability, provider, or cost. Having said that, the main objective of the IOP is to find the best combination of these elements that comply with the user requirements.

In this regard, and for obtaining all the information about the user's needs, the IOP should obtain as input a file written in a modeling language coined as DevSecOps Modelling Language (DOML, [14]). This file includes a specific section in which users introduce their optimization objective and requirements.

For the sake of clarity, we depict in Figure 1 an excerpt of a DOML example. Analyzing this file, we can see how the user wants to find the most optimized configuration, composed of a single ST, one DB and three VMs (depicted in the part `elements`) optimizing three different objectives: cost, availability, and performance. Furthermore, the user deepens on its requirements, introducing three different ones: the cost of the whole deployment should be less than 100€, the overall expected availability must be higher than 98\%, and all the chosen elements should be deployed in Europe.

```
optimization opt {
    objectives {
        "cost" => min
        "availability" => max
        "performance" => max
    }
    nonfunctional_requirements {
        req1 "Cost <= 100.0" max 100.0 => "cost";
        req2 "Availability >= 98.0%" min 98.0 => "availability";
        req3 "Region" values "00EU" => "region";
        req4 "elements" => "Storage, DB, VM, VM, VM";
    }
}
```

Figure 1: An excerpt of a DOML example introduced as input in the IOP.

Considering these objectives and requirements, and resorting the reduced IEC presented above, the IOP codifies candidate solutions as a list of integer values, each one depicting the index of a specific element. Thus, a tentative solution looks like: `[0, 1, 0, 2, 3]: [100, 98.5, 10]`. On the one hand, and deeming that the combination that the user wants to deploy is [ST, DB, VM, VM, VM], the solution `[0, 1, 0, 2, 3]` is equal to [St_Eu, db.sql, C1Italy, m3.small, t4.medium]. On the other hand, `[100, 98.5, 10]` array represents the cost of the whole deployment (100), the expected availability (98.5) and the overall performance (10).

## 3 IAC OPTIMIZER PLATFORM: DESCRIPTION AND ALGORITHMS CONSIDERED

When the IOP receives the input DOML, it follows the following process: *i)* it retrieves the user information from the input file, *ii)* gathers all the available data from the IEC, *iii)* models the optimization problem to solve and *iv)* calculates the optimized IaC deployment configuration, considering the user's objectives and meeting all the introduced requirements.

For carrying out the last of these steps, which is the optimization process, the IOP resorts to a multiobjective solver. As mentioned in the introduction, the prototypical version of the IOP described in [13] employs the well-known NSGA-II for conducting this task. Even so, and despite the fact that the quality of this technique has been proven countless of times in the literature, we must undertake a much more in-depth experimentation to assess if NSGA-II is the most appropriate technique for this context, or if the adoption of another method is more appropriate.

With this motivation in mind, in this paper we measure the performance of nine different EC based multiobjective techniques. Along with the mentioned NSGA-II, the following methods have been deemed:

- *Weighting Achievement Scalarizing Function Genetic Algorithm* (WASFGA, [15]).
- *Global Weighting Achievement Scalarizing Function Genetic Algorithm* (GWASFGA, [16]).
- *Many-Objective Metaheuristic Based on the R2 Indicator* (MOMBI, [17]).
- *Improved Many-Objective Metaheuristic Based on the R2 Indicator* (MOMBI2, [18]).
- *Multiobjective Cellular Genetic Algorithm* (MoCell, [19]).
- *S metric selection evolutionary multi-objective optimization algorithm* (SMSEMOA, [20]).
- *Strength Pareto evolutionary algorithm 2* (SPEA2, [21]).
- *Non-dominated Sorting Genetic Algorithm III* (NSGA-III, [22]).

All these algorithms have been chosen for three main reasons: *i)* they are well-known, and they have been previously validated by the scientific community, *ii)* they can be easily adapted to the discrete problem dealt by the IOP, and *iii)* they can be implemented through the multiobjective framework embraced by the IOP: jMetal JAVA framework [23].

In the following section, we describe the experimentation carried out, detailing how we have measured the performance of each of these methods.

## 4 EXPERIMENTATION

In order to properly assess the performance of all the nine multiobjective solvers, an in-depth experimentation has been carried out using 12 different input DOMLs. Each of these input files has been generated with the intention of emulating real situations that the IOP will face once it is deployed in a real environment. Thus, these DOML files present different objectives to optimize, as well as a variable number of elements to deploy. This directly impacts on the complexity of the problem, depending on the number of objectives and the size of the solution to be found. We summarize in Table 1 the main characteristics of the benchmark composed. As can be seen, each instance has been named as `DOML_A_x-y-z`, where `A` represents the number of objectives, and `x,y,z` the number of VMs, DBs and STs to deploy, respectively. For the sake of replicability, the generated benchmark is openly available under demand, or online at [24].

Table 1: Main characteristics of the 12 input DOMLs generated. Regarding the optimizing objectives: c - cost; a - availability; p - performance. Instances optimizing three objectives consider c & p & a.

| Instance | Objectives | Elements to deploy | | | Requirements |
|---|---|---|---|---|---|
| | | VMs | DBs | STs | |
| `DOML_2_0-4-4` | 2 (C & P) | 0 | 4 | 4 | |
| `DOML_2_1-1-1` | 2 (C & A) | 1 | 1 | 1 | Cost < 200€ |
| `DOML_2_2-2-2` | 2 (C & A) | 2 | 2 | 2 | Availability > 96% |
| `DOML_2_4-3-3` | 2 (C & A) | 4 | 3 | 3 | |
| `DOML_2_5-5-5` | 2 (C & P) | 5 | 5 | 5 | |
| `DOML_2_6-0-0` | 2 (C & A) | 6 | 0 | 0 | |
| `DOML_3_0-4-4` | 3 | 0 | 4 | 4 | |
| `DOML_3_1-1-1` | 3 | 1 | 1 | 1 | Cost < 200€ |
| `DOML_3_2-2-2` | 3 | 2 | 2 | 2 | Availability > 96% |
| `DOML_3_4-3-3` | 3 | 4 | 3 | 3 | |
| `DOML_3_5-5-5` | 3 | 5 | 5 | 5 | |
| `DOML_3_6-0-0` | 3 | 6 | 0 | 0 | |

Furthermore, regarding the IEC that composes the problem, it is comprised of a total of 156 elements. Divided by type, the IEC contains 99 `Virtual Machines`, 24 `Databases` and 33 `Storage` elements. All these elements have realistic features, coming from providers such as Amazon, Google, Openstack or Azure. Arguably, the size is the employed catalog is big enough for reaching rigorous conclusions.

Regarding the parameterization employed for each technique, as recommended in studies such as [25], similar parameters have been employed in all the methods developed. Thus, all methods count on a population composed of 50 individuals, while for the crossover and mutation functions, SBX [26] and Polynomial Mutation [27] have been chosen, respectively. For the selection, Binary Tournament [28] mechanism has been utilized. Finally, regarding the maximum number of evaluations, it has been fixed in 2500.

Having said that, we depict in Table 2 the main results obtained for each technique in all the generated instances. Every instance has been run 10 independent times, in order to obtain statistically relevant results. Thus, we represent in Table 2 the average hypervolume value obtained by each method.

Table 2: Results obtained by each method in all the generated 12 DOML instances.

| Instance | GWASFGA | MoCell | MOMBI | MOMBI2 | SMSEMOA | SPEA2 | WASFGA | NSGA-II | NSGA-III |
|---|---|---|---|---|---|---|---|---|---|
| DOML_2_0-4-4 | 0.725 | 0.891 | 0.976 | 0.817 | 0.966 | 0.858 | 0.632 | **0.998** | 0.907 |
| DOML_2_1-1-1 | 0.790 | 0.964 | **0.989** | 0.597 | 0.947 | 0.816 | 0.523 | 0.966 | 0.784 |
| DOML_2_2-2-2 | 0.796 | **0.983** | 0.973 | 0.923 | 0.922 | 0.982 | 0.825 | **0.983** | 0.973 |
| DOML_2_4-3-3 | 0.937 | 0.964 | 0.976 | 0.937 | 0.966 | **0.985** | 0.973 | 0.964 | 0.913 |
| DOML_2_5-5-5 | 0.546 | 0.928 | 0.890 | 0.824 | 0.961 | 0.939 | 0.670 | **0.970** | 0.950 |
| DOML_2_6-0-0 | 0.886 | 0.980 | 0.924 | 0.901 | 0.966 | 0.983 | 0.805 | 0.995 | **0.998** |
| DOML_3_0-4-4 | 0.626 | 0.977 | 0.969 | 0.749 | 0.969 | 0.953 | 0.497 | 0.987 | **0.996** |
| DOML_3_1-1-1 | 0.718 | 0.544 | 0.660 | 0.648 | 0.745 | 0.693 | 0.610 | 0.727 | **0.750** |
| DOML_3_2-2-2 | 0.543 | 0.727 | 0.711 | 0.592 | 0.734 | 0.709 | **0.841** | 0.625 | 0.756 |
| DOML_3_4-3-3 | 0.549 | 0.858 | 0.963 | 0.674 | 0.970 | 0.965 | 0.590 | 0.957 | **0.966** |
| DOML_3_5-5-5 | 0.513 | 0.971 | 0.790 | 0.822 | 0.944 | 0.927 | 0.517 | 0.946 | **0.968** |
| DOML_3_6-0-0 | 0.458 | 0.940 | 0.975 | 0.526 | **0.984** | 0.972 | 0.610 | 0.964 | 0.944 |

At a first glimpse, we can conclude that the best performing methods along the benchmark are the NSGA-II and the NSGA-III. In any case, and following the guidelines provided in [29], the Friedman's non-parametric test for multiple comparisons [30] has been conducted for confirming these preliminary conclusions. For performing this statistical test, the results average obtained by each method has been used. Thus, we depict in Table 3 the results gathered after performing the test through KEEL platform [31].

Table 3: Average Friendman Rankings for each considered algorithm. The less the ranking, the better the performance.

| Algorithm | Ranking |
|---|---|
| GWASFGA | 7.9583 |
| MoCell | 4.5 |
| MOMBI | 4.1667 |
| MOMBI2 | 7.2083 |
| SMSEMOA | 3.2917 |
| SPEA2 | 4.1667 |
| WASFGA | 7.0833 |
| NSGA-II | **3.1667** |
| NSGA-III | 3.4583 |

Analyzing the results provided by the Friedman's test, we can certify that the best performing technique along the whole benchmark is the NSGA-II, which is the one that has reached the best ranking. In any case, a deeper analysis of the results shown in Table 2 can lead us to much more interesting and valuable conclusions, taking into account that the IOP will work in a real environment in which its performance should be improved as much as possible.

If we analyze the performance of each technique according to the number of objectives it must optimize, we reach to much more accurate insights. Thus, we have conducted two separate Friendman's non-parametric tests. On the one hand, the first test has been performed employing the outcomes obtained for instances `DOML_2_0-4-4`, `DOML_2_1-1-1`, `DOML_2_2-2-2`, `DOML_2_4-3-3`, `DOML_2_5-5-5` and `DOML_2_6-0-0`. On the other hand, results got in use cases `DOML_3_0-4-4`, `DOML_3_1-1-1`, `DOML_3_2-2-2`, `DOML_3_4-3-3`, `DOML_3_5-5-5` and `DOML_3_6-0-0` have been used for performing the second test. In Table 4, we represent the outcomes of these separated Friendman's non-parametric tests.

Table 4: Separated Friendman's test for instances optimizing two and three objectives. Rankings for each considered algorithm. The less the ranking, the better the performance.

| Two Objectives | | Three Objectives | |
| --- | --- | --- | --- |
| Algorithm | Ranking | Algorithm | Ranking |
| GWASFGA | 7.9167 | GWASFGA | 8 |
| MoCell | 4.1667 | MoCell | 4.8333 |
| MOMBI | 3.5833 | MOMBI | 4.75 |
| MOMBI2 | 7.25 | MOMBI2 | 7.1667 |
| SMSEMOA | 4 | SMSEMOA | 2.5833 |
| SPEA2 | 3.6667 | SPEA2 | 4.6667 |
| WASFGA | 7.3333 | WASFGA | 6.8333 |
| NSGA-II | **2.3333** | NSGA-II | 4 |
| NSGA-III | 4.75 | NSGA-III | **2.1667** |

The findings that can be drawn by examining the results shown in Table 4 are even more valuable than those described above. Despite NSGA-II is, in overall, the method that performs better for the whole benchmark, if we undertake a separate analysis we see that NSGA-III behaves better for instances where the number of objectives to optimize is three.

As a final conclusion, and after carrying out the experimentation described in this study, we determine that the most appropriate way for implementing the IOP is to convert it into a multi-method system. Thus, the IOP should have a pool composed of two solving algorithms, using each of them according to the user needs: for problems with two optimizing objectives, NSGA-II should be used; while for situations with three objectives, the IOP should resort to NSGA-III.

## 5 CONCLUSIONS AND FURTHER WORK

This paper has gravitated around a multiobjective problem related to optimizing Infrastructure-as-Code deployment configurations. The system implemented for solving this problem has been coined as *IaC Optimizer Platform* (IOP), and it has been developed in the context of a European H2020 project. Despite a prototypical version of the IOP has been introduced in the literature before [13], a deeper analysis on the multiobjective solving of the problem was needed, in order to determine which is the most appropriate method for embedding in the IOP.

With this motivation in mind, in this work we have tested the performance of nine different multiobjective EC methods through 12 different instances of the problem. Each use case has been run 10 independent times by each considered method and obtained results have been compared resorting to the Friendman's non-parametric test. The main conclusion drawn is that the IOP should be implemented as a multi-algorithm system, using NSGA-II and NSGA-III depending on whether the number of objectives to be optimized is two or three, respectively.

As future work, the further evolution of the IOP has been planned. The deployment of the findings reached on this work should be implemented, and the behavior of the IOP on real-world environments should be analyzed in a deeper manner. On this direction, we have planned to make a deeper experimentation considering aspects such as noisy objectives [32]. As long-term work, the development of more sophisticated algorithms will be implemented for being embedded in the IOP. Other possible future work is related to extending the conducted tests to problems related to other fields in order to have a clear vision of the performance of selected EC multiobjective algorithms in use cases such as the Job-Shop Scheduling Problem [33] or other problems related to economics [34] or energy [35]. Finally, we intend to deal with the same problem resorting to revolutionary paradigms such as quantum computing [36], [37].

## ACKNOWLEDGMENTS

This research has received funding from the European Union's Horizon 2020 research and innovation programme under grant agreement No: 101000162 (PIACERE project).